\documentclass{amsart}        

\usepackage{amssymb,latexsym}
\usepackage{graphicx}
\usepackage{subfig}
\usepackage{algorithmic}

\newenvironment{mylisting}
{\begin{list}{}{\setlength{\leftmargin}{1em}}\item\scriptsize\bfseries}
{\end{list}}

\begin{document}
\title{Implementing Genetic Algorithms on Arduino Micro-Controllers}
\author{
  Nuno Alves \\
  nuno@brown.edu \\
  Engineering Division, Brown University
}

\begin{abstract}
Since their conception in 1975, Genetic Algorithms have been an extremely popular approach to find exact or approximate solutions to optimization and search problems. Over the last years there has been an enhanced interest in the field with related techniques, such as grammatical evolution, being developed. Unfortunately, work on developing genetic optimizations for low-end embedded architectures hasn't embraced the same enthusiasm. This short paper tackles that situation by demonstrating how genetic algorithms can be implemented in Arduino Duemilanove, a 16 MHz open-source micro-controller, with limited computation power and storage resources. As part of this short paper, the libraries used in this implementation are released into the public domain under a GPL license.
\end{abstract}
\maketitle



\vspace*{-2mm}
\section{Introduction}\label{sec:introduction}

In the 1950s, while studying industrial organizations, Herbert Simon pointed out that when confronted with real-world problems, human beings have a tremendous difficulty with maximizing the outcomes. His rationale is that we lack the cognitive abilities required to evaluate all outcomes with sufficient precision, with our weak and unreliable memories. He stated that when present with a particular problem, we are forced to make decisions not by maximization but by \emph{satisficing}~\cite{adams69}. This \emph{satisficing} criterion consists of setting a particular  aspiration level which, if achieved, they will be happy enough with, and if they don't, try to change either their aspiration level or their decision.  A Genetic Algorithm (GA), is a search space algorithm, which follows this \emph{satisficing} criterium. In most of the optimization problems, often times it is more important to obtain a solution that meets our requirements than to obtain an optimal solution. In other words, when confronted with a problem, we are looking for some good , \emph{satisficing} level of performance quickly with reasonable resources.

\subsection{Genetic Algorithms}
A genetic algorithm is a technique used in computing to find exact or approximate solutions to optimization and search problems. Genetic algorithms are a particular class of evolutionary algorithms (EA) which uses techniques inspired by evolutionary biology such as inheritance, mutation, selection, and crossover. Genetic algorithms are implemented in a computer simulation in which a population of many individuals, each containing a potential solution to the proposed  problem. Traditionally, these solutions are represented in binary as strings of 0s and 1s, appropriately named chromosomes. The GA algorithm starts with a population of randomly generated individuals, and in each generation, the fitness of every individual in the population is evaluated, with multiple individuals  stochastically selected from the current population (based on their fitness), and modified (recombined and possibly randomly mutated) to form a new population. The new population is then used in the next iteration of the algorithm. The algorithm terminates when the maximum number of iterations have been processed or when  a satisfactory fitness level has been reached for the population. Clearly, if the algorithm has terminated due to a maximum number of generations, a satisfactory solution may not have been reached. An introductory reference to GAs can be found in ~\cite{GA89,citeulike:1449453}.

Genetic algorithms are suited to a variety of search problems, provided one he can encode solutions of a given problem into a binary string (chromosomes) and evaluate the quality of that particular solution. The lasting appeal of GAs is brought from its simplicity and ability to quickly discover solutions for difficult high-dimensional problems. GAs have shown to be useful when:
\begin{itemize}
\item Search space is large, complex or poorly understood.
\item Domain knowledge is scarce or expert knowledge is difficult to encode to narrow the search space.
\item No mathematical analysis is available.
\item Traditional search methods fail.
\end{itemize}

Genetic Algorithms have been used extensively in many diverse areas on different optimization problems, such as the study of social systems and cooperations ~\cite{Mitchell93geneticalgorithms}, numerical optimizations ~\cite{Michalewicz95geneticalgorithms}, VLSI cell area optimization ~\cite{4530477}, quantum circuit design ~\cite{Yabuki00geneticalgorithms}, synthesis of QCA circuits ~\cite{1191717}, index fund portfolio selection ~\cite{970454}, job scheduling ~\cite{5366286} and antenna design ~\cite{1217422}. In this paper we release an implementation of GAs in the Arduino Duemilanove (ATmega328 - rev 2009b), a low-end micro-controller.

\subsection{Arduino Duemilanove}
Arduino Duemilanove  is an open-source electronics prototyping platform consisting of an 8-bit Atmel AVR micro-controller clocked at 16 MHz. It is powered at 5V and contains 32 KB of flash memory contents of which 2 KB used by boot-loader. These specifications should be enough to demonstrate the limitations of this particular micro-controller.
We decided to implement a genetic algorithm on an arduino micro-controller for the following reasons:

\begin{itemize}
\item Inexpensive and easily obtainable: on February 2020 the ATmega328 unit price was \$4.30.
\item Open source: the IDE software and the hardware reference files have both been released under the GPL open-source license.
\item Development Ease: the core implementation language closely resembles C, and its graphical IDE is freely available on multiple popular operating systems.  
\item Established user base: the arduino platform has been used is many introductory embedded systems university-level classes.
\item Low power requirements: On standby mode the arduino micro-controller requires a current of 0.2 mA. If an average arduino application requires around 100mA, that would entail the power consumption to be on the proximity of 0.5 watt.
\end{itemize}

This GA implementation will also serve as a proof of concept in which we demonstrate that even extremely limited hardware can still be used in solving practical real world problems.

\section{Implementation}
Section~\ref{sec:sourcecode} contains the source code for the Arduino GA library, \emph{GA.cpp} and \emph{GA.h}, and a sample optimization problem, \emph{main.cpp}. To run this code, one must copy the \emph{main.cpp} source into the active Arduino project, and following the instructions on the Arduino support documentation~\cite{arduinolibrary}, include \emph{GA.cpp} and \emph{GA.h} as a single library. By running \emph{main.cpp} and selecting the \emph{Serial Monitor} option, the outcome of the sample optimization problem will be displayed. 

\subsection{Sample optimization problem}
The main idea behind a GA is that we have a particular problem we want to know a possible solution for it. We do not know the optimal solution \emph{a priori}, but we can tell how good a particular solution is. A GA run starts with a certain number of individuals with all its chromosomes set to a random binary state. These random bits can be encoded into a solution for the given problem. Our sample optimization problem consists on maximizing the function $f(x)=x$ on the integer interval [0,31]. In other words, we want our best individual to have as many bits set to logic one as possible. Each time a new iteration is processed, each individual will have a modified set of chromosomes, which will be evaluated by our fitness function. This fitness function evaluator is coded inside \emph{main.cpp}. This fitness function can be adjusted to solve any particular problem. In fact, the main difficulty with GAs is the encoding of a solution in with a binary state. Example of binary encodings for various problems can be found in ~\cite{citeulike:1449453} 

\subsection{Limitations and Discussion}
This GA implementation contains just a simple bare-bones, single crossover point as described in~\cite{GA89}. Due to hardware constraints, these algorithms have some limitations. 
\begin{itemize}
\item The population size is limited to 100 individuals. This is user specified in \emph{main.cpp}. 
\item Each individual has a hardwired 32 bit chromosomes. This means that the solution encoding is constrained by those number of bits. 
\item The value of each fitness evaluation variable, which states how good a particular solution is, must be between 0 and 100.
\end{itemize}

\section{Conclusion}
Our unbridled technological lust is constantly pressuring the development of newer, faster and smaller micro-controllers. In our efforts to develop algorithms to high-end, and expensive micro-controllers, lower-end devices are being neglected. This particular work described an implementation of a genetic algorithms customized for the Arduino micro-controller. In here we demonstrated that the Arduino micro-controller embedded system can successful be used to solve search space problems at a low cost. Future work will consist of investigating practical engineering uses where GA implementations on Arduino micro-controllers can have a distinct impact, by taking advantage of its open-source nature, low cost and low power requirements. This may involve parallelizing this implementation across multiple Arduino micro-controllers. This approach will involve storing information in a single arduino board, and using it to dynamically program connected \emph{blank} micro-controllers. Further work also needs to be performed as to assert the practicality of using Arduino micro-controllers to solve complex engineering problems regarding different metrics, such as computation time, power consumption and cost per clock cycle. 

\section{Source code}\label{sec:sourcecode}
The source code for this GA arduino library has now been released into the public domain under a GPL license and can be directly copied from this paper, or downloaded from \emph{http://www.nunoalves.com/source/GA-revA.zip}.

\subsection{main.cpp}

%

\begin{mylisting}
\begin{verbatim} 

//main.cpp : Sample arduino entry point that uses the arduinoGA library.
//Copyright (C) 2010 Nuno Alves
     
#include <GA.h>

void setup()
{
     Serial.begin(9600);
     randomSeed(analogRead(0));
}

int populationsize=99;   //select size in the range [0..99]
int numgenerations=100;  //number of generations cannot exceed 65k
//the unit of bit mutation is per thousand. For example, if we set the
//bitmutation to 10, approximately every 10 out of 1k bits will be mutated
int bitmutation=1;       

//initializing the arduinoGA library class. 
//this will initialize a set of random genes.
GA ga(populationsize,numgenerations,bitmutation);

//we start in the generation 0
int generation=0;

void loop()
{

     if (generation==0) { Serial.println(""); Serial.println(""); }

     while (generation<numgenerations)
     { 

          //======================================================================
          //Fitness Evaluation
          //======================================================================
          //This is the part where we "how good" a particular combination of
          //genes is. In this simple case, we just want to maximize the number of 1's 
          //in the genes. The 1's can be for example, the number of lit LED ligths.
          //NOTE: due to memory constraints, the fitness value CANNOT(!!!) be greater
          //than 100.
          //======================================================================
          //begin: Fitness Evaluation
          for (int i=0 ; i < populationsize ; i++)
          {
               unsigned int t0_fitness=0;
               unsigned int t0_a_population=ga.read_t0_a_population(i); 
               unsigned int t0_b_population=ga.read_t0_b_population(i); 

               ///we are evaluating the fitness on 32 bits... dividing by
               //ga.t0_a_population[i] and ga.t0_b_population[i]
               for (int j=0 ; j < 16 ; j++)
               {
                    if(bitRead(t0_a_population, j)==1)
                    {
                         t0_fitness=t0_fitness+1;
                    }

                    if(bitRead(t0_b_population, j)==1)
                    {
                         t0_fitness=t0_fitness+1;
                    }
               }
               ga.write_t0_fitness(i,t0_fitness);
          }
          //end: Fitness Evaluation

          //the next steps perform the genetic optimization process
          //behind the scenes. It mates the best individuals and create the 
          //next generation with a new set of genes.   

          ga.process_generation(generation);
          //change the reportStatistics value from 0 to 3 for more detailed printout
          ga.reportStatistics(generation,0);
          ga.prepare_next_generation();
          generation=generation+1;
     
     } //while (generation<numgenerations)     

}//end of void loop()



\end{verbatim}
\end{mylisting}

\subsection{GA.h}
\begin{mylisting}
\begin{verbatim} 

//GA.h : header file for the arduinoGA library.
//Copyright (C) 2010 Nuno Alves

#ifndef GA_h
#define GA_h

#include "WProgram.h"
#define GPOPSIZE 100

class GA
{
     public:
          GA(int _popsize, int _ngenerations, int _bitmutation);
          void process_generation(int generationid);
          void reportStatistics(int generationid, int verbosityLevel);
          void prepare_next_generation();

          void write_t0_fitness(int idx, unsigned int value) { t0_fitness[idx]=value; }
          unsigned int read_t0_a_population(int idx) { return(t0_a_population[idx]); }
          unsigned int read_t0_b_population(int idx) { return(t0_b_population[idx]); }

     private:
          unsigned int t0_fitness[GPOPSIZE];
          unsigned int top_fitness_val;
          unsigned int sum_fitness_val;
          unsigned int t1_a_population[GPOPSIZE];
          unsigned int t1_b_population[GPOPSIZE];
          unsigned int bestcandidate_a;
          unsigned int bestcandidate_b;
          unsigned int t0_a_population[GPOPSIZE];
          unsigned int t0_b_population[GPOPSIZE];
          unsigned int next_gen_expected_count[GPOPSIZE];

          void randomize_t0_population();
          void evaluate_t0_fitness();
          void expected_count_t1();
          void populate_t1();
          void mate_t1();

          int bitmutation; //should be within the range [0..1000]
          int popsize;
          int ngenerations;
};

#endif

\end{verbatim}
\end{mylisting}

\subsection{GA.cpp}
\begin{mylisting}
\begin{verbatim} 

//GA.cpp : arduinoGA library. This file contains the function implementations
//Copyright (C) 2010 Nuno Alves

#include "WProgram.h"
#include "GA.h"

GA::GA(int _popsize, int _ngenerations, int _bitmutation)
{
     popsize      = _popsize;
     ngenerations = _ngenerations;
     bitmutation  = _bitmutation;
     
     //initialize the entire population
     randomize_t0_population();
}

//verbosityLevel=0,1,2
void GA::reportStatistics(int generationid, int verbosityLevel)
{

     if (verbosityLevel>=0)
     {
          Serial.print("generation=");
          Serial.print(generationid);
          Serial.print(" , ");
          Serial.print("top fitness=");
          Serial.print(top_fitness_val);
          Serial.print(" , ");
          Serial.print("sum fitness=");
          Serial.println(sum_fitness_val);
     }

     if (verbosityLevel>=1)
     {
          Serial.print("top candidate (BIN):");
          Serial.print(bestcandidate_a,BIN);
          Serial.print(" ");
          Serial.println(bestcandidate_b,BIN);

          Serial.print("top candidate (DEC):");
          Serial.print(bestcandidate_a,DEC);
          Serial.print(" ");
          Serial.println(bestcandidate_b,DEC);
     }

     if (verbosityLevel>=2)
     {
          Serial.println("printing each individual+fitness");
          for (int i=0 ; i < popsize ; i++)
          {
               Serial.print(t0_a_population[i],BIN);
               Serial.print(" ");
               Serial.print(t0_b_population[i],BIN);
               Serial.print(" (");
               Serial.print(t0_fitness[i]);
               Serial.println(")");
          }

          Serial.println("printing next gen expected count");
          for (int i=0 ; i < popsize ; i++)
          {
               Serial.print("individual=");
               Serial.print(i);
               Serial.print(" , ");
               Serial.println(next_gen_expected_count[i]);
          }

          Serial.println("printing each of next gen individuals");
          for (int i=0 ; i < popsize ; i++)
          {
               Serial.print(t1_a_population[i],BIN);
               Serial.print(" ");
               Serial.println(t1_b_population[i],BIN);
          }

     }     
}

//processes all the steps for current generation
void GA::process_generation(int generationid)
{
     //calculates the average and find the best individual
     evaluate_t0_fitness();

     //whats the probability of a particular individual
     //show up on t1?
     expected_count_t1();

     //populate the next generation array with the best performing individuals
     populate_t1();

     //print some pre-mate statistics
     //report_pre_mate_statistics(generationid);

     //mate every two elements at a random crossover point
     mate_t1();
}

//randomize all genes of each individual in the t0_population
void GA::randomize_t0_population()
{
     for (int i=0 ; i < popsize ; i++)
     {
          unsigned int randNumberA = random(); 
          t0_a_population[i]=randNumberA;
     }
     for (int i=0 ; i < popsize ; i++)
     {
          unsigned int randNumberA = random(); 
          t0_b_population[i]=randNumberA;
     }
}

//store just the current generation fitness to save memory
void GA::evaluate_t0_fitness()
{
     top_fitness_val=0;
     sum_fitness_val=0;

     for (int i=0 ; i < popsize ; i++)
     {
          sum_fitness_val = t0_fitness[i]+sum_fitness_val;

          if (t0_fitness[i]>top_fitness_val) 
          {
               top_fitness_val=t0_fitness[i];     
               //begin: store the best candidate for viewing
               bestcandidate_a=t0_a_population[i];
               bestcandidate_b=t0_b_population[i];
               //end: store the best candidate for viewing
          }
     }
}

void GA::expected_count_t1()
{   
     //sort all elements in ascending order
     //based on the fitness stored in t0_fitness[i]...

     //begin: bubble sort the array
     for (int x=0; x < popsize-1; x++)
     {
          for (int y=0; y < popsize-x-1; y++)
          {
               if (t0_fitness[y] > t0_fitness[y+1])
               {
                    unsigned int t=t0_fitness[y];
                    unsigned int A=t0_a_population[y];
                    unsigned int B=t0_b_population[y];

                    t0_fitness[y]=t0_fitness[y+1];
                    t0_fitness[y+1]=t;

                    t0_a_population[y]=t0_a_population[y+1];
                    t0_a_population[y+1]=A;

                    t0_b_population[y]=t0_b_population[y+1];
                    t0_b_population[y+1]=B;
               }
          }
     }
     //end: bubble sort the array           

     //since we limited each fitness element not to surpass 100
     //by letting 100 population size we can have a final sum of 10k.
     //however the unsigned integer can hold at least the number 65,535.
     //this means we could theoretically have a population size of 650. 

     //next_gen_expected_count[] will have a
     //cdf on the  next_gen_expected_count[] array
     unsigned int current_value=0;
     for (int i=0 ; i < popsize ; i++)
     {
          current_value=current_value+t0_fitness[i];     
          next_gen_expected_count[i]=current_value;    
     }
}

void GA::populate_t1()
{

     for (int i=0 ; i < popsize ; i++)
     {
          unsigned int randomValue=random(0,next_gen_expected_count[popsize-1]);    
          int select_index=0;

          for (int j=1 ; j < popsize ; j++)
          {
               if ((randomValue>=next_gen_expected_count[j-1]) && (randomValue<next_gen_expected_count[j]))
               {
                    select_index=j; 
                    j=popsize; //exit the loop
               }  
          }

          t1_a_population[i]= t0_a_population[select_index];
          t1_b_population[i]= t0_b_population[select_index];
     }

}

//switch bits between two individuals at a particular cross point
void GA::mate_t1()
{

     for (int i=0 ; i < popsize ; i=i+2)
     {
          int crossoverSite=random(0,31);    
          int crossover_siteA;
          int crossover_siteB;

          //begin: mutate genes a bit. For simplicity we just randomize an entire
          //set of genes...
          unsigned int randomMutationA=random(1,1000);
          unsigned int randomMutationB=random(1,1000);
          if (randomMutationA<=bitmutation){ t1_b_population[i] = random(); }
          if (randomMutationB<=bitmutation){ t1_b_population[i] = random(); }
          //end: mutate genes a bit

          if (crossoverSite<16)
          {
               crossover_siteA=crossoverSite; //keep values on array A till crossoversite
               crossover_siteB=0; //cross all values on array B
          }
          else
          {
               crossover_siteA=16; //dont mate any values of array A     
               crossover_siteB=32-crossoverSite;
          }


          //mate array A
          for (int j=crossover_siteA ; j < 16 ; j++)
          {          
               bool siteA=bitRead(t1_a_population[i],j);
               bool siteB=bitRead(t1_a_population[i+1],j);
               bitWrite(t1_a_population[i],j,siteB);
               bitWrite(t1_a_population[i+1],j,siteA);     
          }

          //mate array B
          for (int j=crossover_siteB ; j < 16 ; j++)
          {
               bool siteA=bitRead(t1_b_population[i],j);
               bool siteB=bitRead(t1_b_population[i+1],j);

               bitWrite(t1_b_population[i],j,siteB);
               bitWrite(t1_b_population[i+1],j,siteA);     
          }

     } //for (int i=0 ; i < popsize ; i=i+2)

}

//this function just copies each element of t1 into t0
void GA::prepare_next_generation()
{
     for (int i=0 ; i < popsize ; i++)
     {
          t0_a_population[i]=t1_a_population[i];
          t0_b_population[i]=t1_b_population[i];
     } 
}

\end{verbatim}
\end{mylisting}

\bibliography{paper}
\bibliographystyle{abbrv}

\end{document}